\def\year{2020}\relax
\documentclass[letterpaper]{article} 
\usepackage{aaai20}  
\usepackage{times}  
\usepackage{helvet} 
\usepackage{courier}  
\usepackage[hyphens]{url}  
\usepackage{graphicx} 
\usepackage{graphicx}  
\nocopyright
\usepackage{algorithm}
\usepackage{algpseudocode}
\usepackage{amssymb}
\usepackage{amsmath}
\usepackage{xcolor}
\usepackage{listings}
\usepackage{multirow}
\usepackage{pifont}

\newtheorem{definition}{Definition}

\usepackage{float}

\floatstyle{ruled}
\newfloat{agent}{thp}{lop}
\floatname{agent}{Agent}

\title{Investigating Human Response, Behaviour, and Preference in Joint-Task Interaction}
\author{Alan Lindsay,\textsuperscript{\rm 1} Bart Craenen,\textsuperscript{\rm 1} Sara Dalzel-Job,\textsuperscript{\rm 2} Robin L. Hill,\textsuperscript{\rm 2} Ronald P. A. Petrick\textsuperscript{\rm 1}\\
\textsuperscript{\rm 1} Department of Computer Science,  Heriot-Watt University, Edinburgh, Scotland, UK\\
\textsuperscript{\rm 2} School of Informatics, University of Edinburgh, Edinburgh, Scotland, UK\\
\{alan.lindsay,b.craenen,r.petrick\}@hw.ac.uk, \{sdalzel,r.l.hill\}@ed.ac.uk}

\begin{document}

\maketitle

\begin{abstract}
Human interaction relies on a wide range of signals, including non-verbal cues.
In order to develop effective Explainable Planning (XAIP) agents it is important that we understand the range and utility of these communication channels.
Our starting point is existing results from joint task interaction and their study in cognitive science.
Our intention is that these lessons can inform the design of interaction agents---including those using planning techniques---whose behaviour is conditioned on the user's response, including affective measures of the user (i.e., explicitly incorporating the user's affective state within the planning model).
We have identified several concepts at the intersection of plan-based agent behaviour and joint task interaction and have used these to design two agents: one reactive and the other partially predictive. 
We have designed an experiment in order to examine human behaviour and response as they interact with these agents.
In this paper we present the designed study and the key questions that are being investigated.
We also present the results from an empirical analysis where we examined the behaviour of the two agents for simulated users.
\end{abstract}

\section{Introduction}
A common approach to explanation generation in automated planning has been to treat the problem as one of model reconciliation~\cite{chakraborti2017plan}. In this way, a balance can be made between generating explanations that update the user's model of the environment and selecting explicable action sequences, which need no further explanation. At the heart of this approach is an accurate user model, which is not practical in many applications.
An alternative view is to see explanations within the wider context of interaction. 

Humans adopt a wide range of communicative channels during interaction, producing both conscious and subconscious responses, as well as using a variety of behavioural heuristics.
In order to communicate effectively an interaction agent must be able to detect and interpret these signals or, preferably, anticipate them.
With this context in mind, we are preparing a user study to observe human behaviour and response in a joint task interaction with a plan-based agent.
While we believe this study should confirm and extend previous research that has explored joint task interaction, our main focus is on discovering how concepts related with plan-based agents (e.g., plan generation, agent intention) can be utilised to inform the agent's behaviour.

We have identified plan-based agent behaviours that interact with key aspects of joint task interaction (e.g., uncertainty and knowledge differences).
These behaviours have been used to define two alternative agents: a reactive agent, which represents an initial interaction, e.g., where little is known about the human's preferences and possibly the environment; and a partially predictive agent, which represents a more informed agent that provides information in advance and attempts to proactively avoid user uncertainty.
By gathering both objective and subjective data from humans interacting with these agents, we hope to gain a better understanding of how such agents could be designed to be more acceptable to humans.


As part of this work, we have developed a web-based user study that we are currently preparing to deploy.
The website supports a joint task interaction, where the human user must complete a task with the help of an instruction/explanation giving virtual agent.
In this paper, we present the intentions of our user study, the situations that the participants will encounter and the questions we are investigating.
We also present the agents that will be compared in the study and the aspects of joint task interaction that they allow us to examine.
Although we do not yet have results from users, we present an empirical analysis to compare the strategies and explanations generated by each of the agent types. 

We first present an overview of the joint task setting and the necessary planning background. We present aspects of plan-based agent behaviour that are relevant to joint task interactions and then use these to define two alternative agents. We present our intended user study, an empirical analysis, related work and finally conclude.


\begin{figure}
    \centering
    \resizebox{0.3\textwidth}{!}{
    \includegraphics{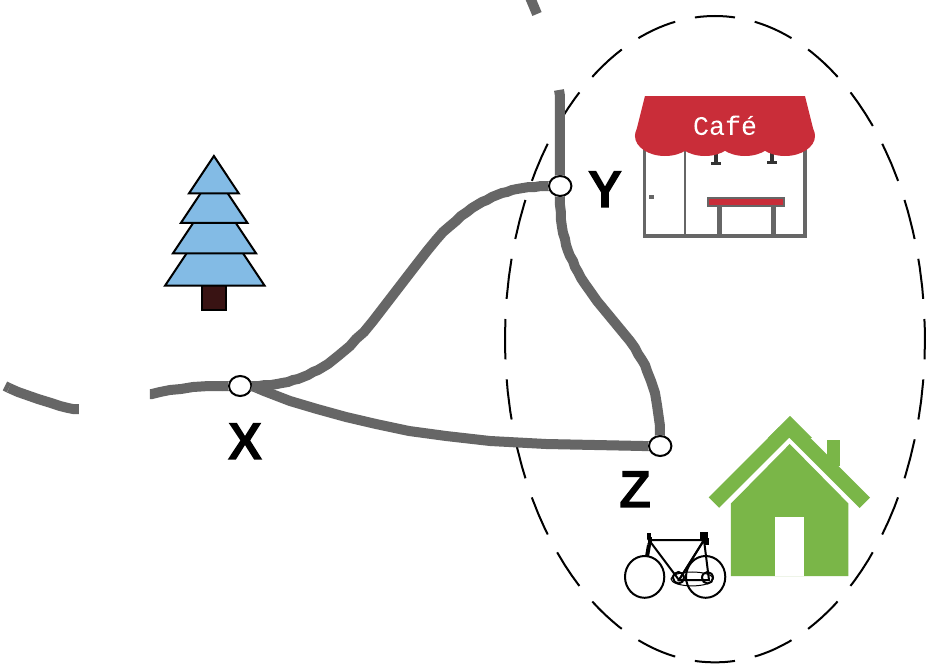}}
    \caption{An example scenario starting at the tree ($X$), with $2$ possible bike locations ($Y$ or $Z$) indicated by a dotted line.}
    \label{fig:ex2choice}
\end{figure}



\section{Interaction in a Joint Task}

We are interested in investigating the range of human social signals, affective responses and behavioural patterns exhibited during co-operative joint action in a shared audio-visual environment. 
In this work, we consider an instruction giving and following scenario, which has been designed based on the HCRC Map Task ~\cite{anderson1991hcrc}\footnote{\url{http://groups.inf.ed.ac.uk/maptask/}} 
and the GIVE Challenge~\cite{byron2009report}.
In the Map Task, an instruction giver guides an instruction follower around a map using landmarks, while in the GIVE challenge the instructions are generated by a computer. The former task was designed to investigate human linguistic behaviour while the latter was proposed as a challenge for testing approaches to natural language generation.
Our task has been designed to enable human-agent interactions to be observed in various scenarios typical of joint-action tasks, such as uncertainty as to an instruction's intent and knowledge differences between the instruction giver and an instruction follower.
These scenarios are not presented to the user in isolation, but as part of a complete solution. 
We are therefore also able to observe how the user responds to the strategy used and how their responses vary as the task progresses.
By situating the interaction within a task, the interaction is focused towards a common goal and can therefore provide  more naturalistic behaviour and feedback on task-based interactions.

In this work, we underpin the agent's decisions and behaviour with a planning model that is used both to generate the agent's strategy, as well as construct situation-based instructions and explanations. 
Our intention is to investigate the relationships that exist between the information that the listener has been presented and their subsequent responses within specific situations.
For example, how does the user respond to an ambiguous instruction and is their response different if they already have an idea of the agent's intentions (e.g., they believe they are heading north)?
Understanding the human listener's needs and preferences during interaction can inform our design of virtual agents that must manage these requirements as they arise during the interaction.


\subsection{The Bike Sharing Task}
Our scenario involves a simple bike share company, which receives partial reports about the location of bikes.
The virtual agent provides instructions to guide the human user around a map as they locate and collect the bikes. Figure~\ref{fig:ex2choice} presents an example map, with the user currently located at position $X$ and the possible locations of a bike indicated by a dotted line. 
The map identifies a collection of landmarks, which are each illustrated with a recognisable landmark (e.g., a green house or blue tree). 
The landmarks are organised into several districts (e.g., the Northern District).


\subsection{The Interaction Agent}

We assume a virtual instruction giver (agent) is used to communicate with the listener, providing instructions to assist them in the task, e.g., the agent might instruct the user to `Go to the house' in the example in Figure~\ref{fig:ex2choice}.
The agent selects instructions based on a strategy, which is updated depending on the listener's actions (e.g., if they deviate from the intended path or fail to act).
The agent also provides explanations to the listener so that they can better understand the situation and the intention behind the agent's strategy.
The agent's strategy is underpinned by a partially-observable planning model.

\subsubsection{The Agent's Planning Model}
A partially observable planning problem, e.g.,~\cite{bonet2011planning}, can be defined by a tuple, $P=\langle F,A,M,I,I^P,G\rangle$, with fluents $F$, actions $A$, sensor model $M$, the actual initial state $I$, the positive and negative literals of the state known by the agent $I^P \subseteq I$, and goal $G$.
An action is defined by its preconditions and effects. An action is applicable if its preconditions are satisfied in the agent's partial state and the application of an action causes its effects to be applied to the agent's current state.
Sensing actions are triggered whenever they become applicable and their observations are applied to the agent's state.
The set of potential agent (partial) states is represented by $S^P$ and can be enumerated through expansion from the agent's initial state.
A solution to the problem is a state-action policy, $\pi:S^P\mapsto A$, such that a simple executive can use the policy to iteratively step from the initial state to a state that satisfies the goal (by looking up the current state in the policy mapping and applying the selected action). 
We will use $\pi^<$, to denote the previously executed sequence of actions and $\pi^>$ to denote the agent's intended action sequence.\footnote{Although this might be a branched plan none of the analysis in this work will extend beyond any branching points.}

The agent's planning model captures a basic transportation style domain, supporting traversal between landmarks and a pickup bike action. 
The bike locations are initially unknown. The partial reports (such as `bike 1 is in the Western District') are used to partially constrain the possible initial states. Sensing actions for a particular bike and location pair determine whether a bike is at the location and are activated when the agent is at the location.

\paragraph{Instruction Generation}
The model is used to generate the agent's strategy and their strategy is used at each step to generate the next instruction.
For example, in the scenario illustrated in Figure~\ref{fig:ex2choice}, the next plan action might be (\texttt{move tree green\_house}). This action is linked to a specific speech utterance, such as `Go to the green house.'
Controlling the information presented to the user at each stage of the interaction provides a key opportunity for generating the types of scenarios that we are interested in examining (see the following section). 

\section{Plan-Based Agents within Joint Tasks}


We have identified plan-based agent behaviours (e.g., generating explicable plans and plan summarisation) that interact with aspects of joint-task interaction (e.g., uncertainty and knowledge difference).
The intention is to use these behaviours to parameterise the interaction agent (see Section `The Agents') 
allowing us to experiment with the interaction under different conditions.
In each condition we can observe human response, behaviour and preferences within various scenarios common in joint-task interaction.
Through these observations we hope to gain insights that will help in the future design of human acceptable agents.
In the remainder of this section, we define the plan-based agent behaviours and describe their links with joint-task interactions.

\begin{figure}
    \centering
    \resizebox{0.28\textwidth}{!}{
    \includegraphics{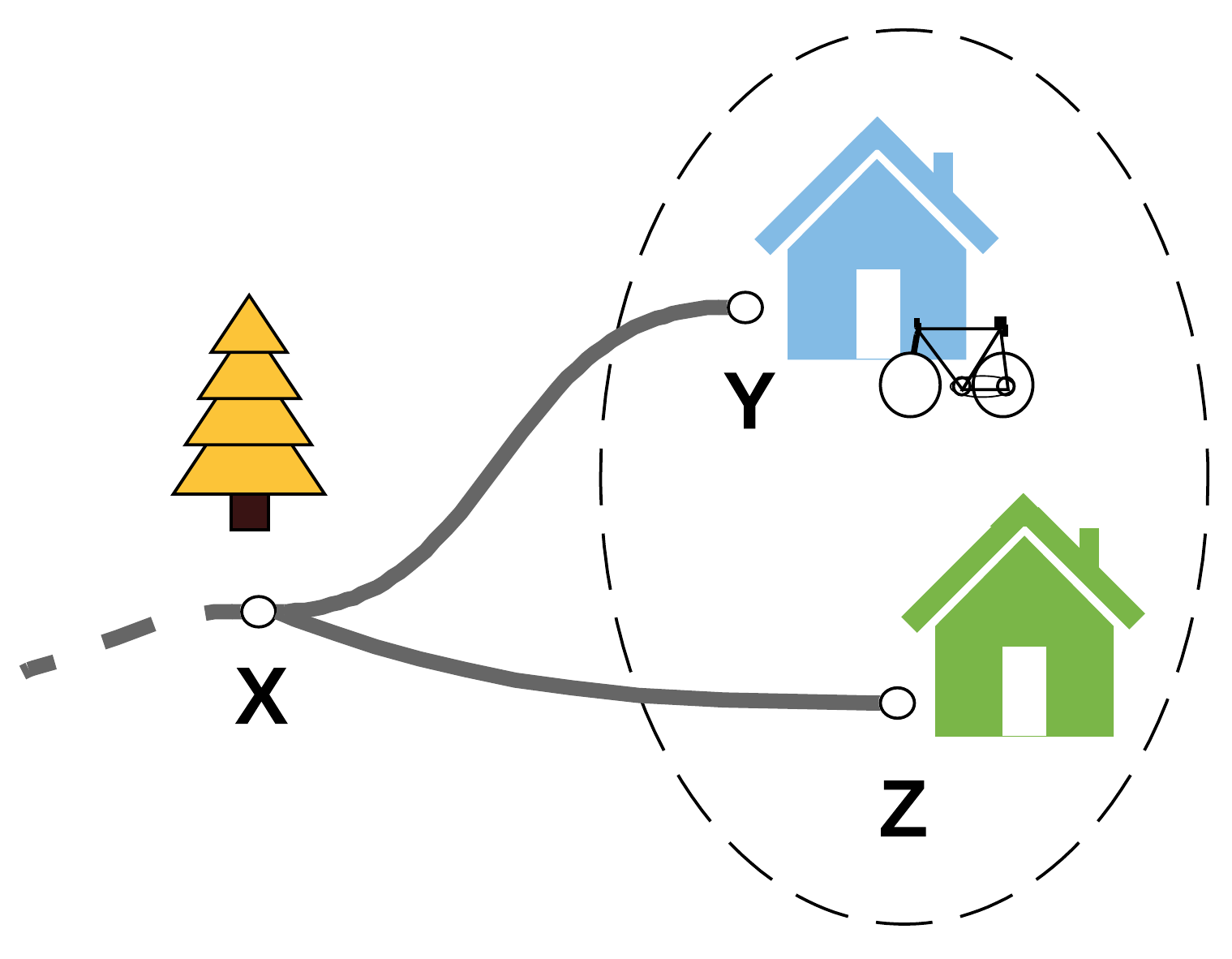}}
    \caption{A scenario where $X$ is connected to two houses.}
    \label{fig:amb}
\end{figure}

\begin{table}
\centering
\begin{tabular}{r|l}
\bf Sys & `Move to the House'\\
\bf User & [Dithering] `Um'\\
\bf Sys & `The House with the Blue roof?'\\
\bf User & `Ah, yes.' $\langle$ Moves to House $\rangle$\\
\end{tabular}
\caption{\label{tab:dither} Example dialogue for the Bike Share task. The user is initially not sure, but on further elaboration understands and enacts the instruction.}
\end{table}

\subsection{Uncertainty in Instructions}
A common aspect of interaction is unclear or ambiguous instructions, which can be  
caused by under-explaining or different viewpoints.
Uncertainty was a focus of the original HCRC Map Task~\cite{anderson1991hcrc}, which has been widely used for analysing and understanding human dialogue within a collaborative joint task context.

We are interested in understanding how the user responds to situations where they are uncertain about the intention of the instruction.
Consider the example interaction presented in Table~\ref{tab:dither}, in the context of position $X$ in Figure~\ref{fig:amb}. The agent instructs the listener to go to the house, but there are two possible houses. The listener is therefore hesitant.
On detecting this hesitation the agent further elaborates on the instruction.
These situations allow us to analyse listener responses, both in terms of verbal replies and affective measures such as frowning and rapid eye movement (e.g., between alternatives), as well as understanding better any default heuristics that humans adopt in uncertain situations.

These situations are created in our experiment as a combination of the map design, which incorporate situations similar to the example in Figure~\ref{fig:amb}, and a policy of under-explaining, similar to the dialogue presented in Table~\ref{tab:dither}.
In these situations the agent can then elaborate on the instruction, providing the necessary information to disambiguate.

\begin{figure}
    \centering
    \resizebox{0.3\textwidth}{!}{
    \includegraphics{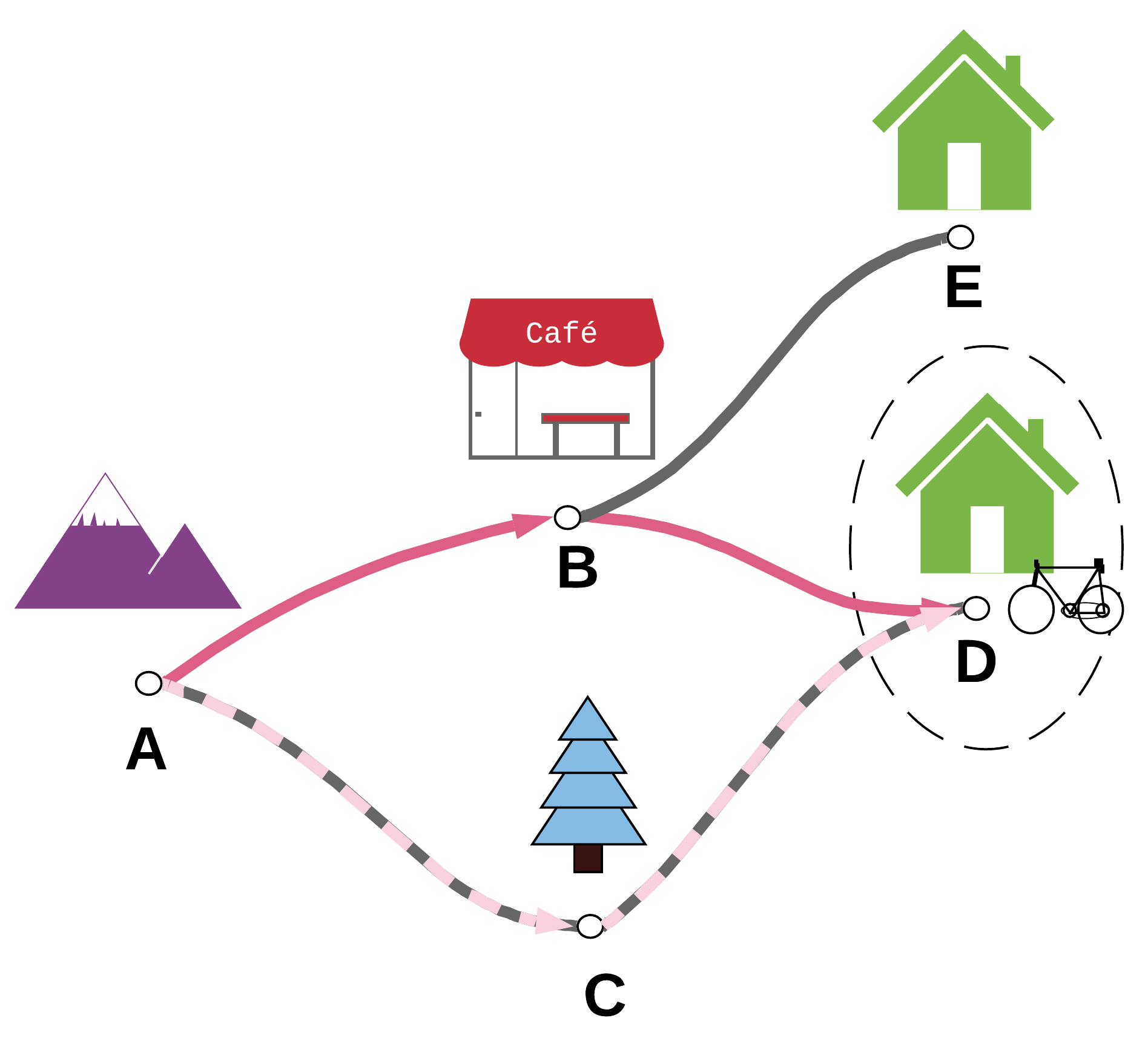}}
    \caption{The agent at $A$ generates a plan to collect the bike at $D$. The solid pink line illustrates a typical plan, which includes the ambiguous instruction moving from $B$ to $D$. The dotted line illustrates a plan found when the planner is aware of ambiguous instructions.}
    \label{fig:premon}
\end{figure}

\subsection{Explicable Plans}

Explicable plans require fewer explanations as they coincide more closely with the user's expectations~\cite{chakraborti2017plan}.
\begin{definition}
Explicability is defined as an ordering relationship, $<_E$, over plan sequences:
Given a user model as a function $U:\Pi \mapsto \mathbb{I}$, mapping plans to inexplicability scores (e.g., the number of explanations required by the user for the plan to make sense) and two plan traces: $\pi^0$ and $\pi^1$ then $\pi^0<_E\pi^1$ ($\pi^0$ is less explicable than $\pi^1$) if 
$U(\pi^0) > U(\pi^1)$.
\end{definition}
The issue of generating explicable plans has been considered within a model reconciliation framework~\cite{chakraborti2017plan}, but requires an accurate user model.
We consider it more likely that understanding how users respond during an interaction will be a gradual process.
Therefore, instead of requiring a complete user model, we should be able to exploit partial information (that ideally becomes more certain during the interaction).
We present an approach to extend the agent's model with additional knowledge about the interaction, allowing the agent to proactively shape their strategy to make their plan more explicable.

We have focused on the uncertainty of ambiguous instructions. 
For example, after a period of time it may become clear that certain instructions, e.g., $a$ and $b$, are commonly mistaken (e.g., frequently the listener goes the wrong way or appears hesitant).
In this work, we deliberately place the user in uncertain situations and we can therefore directly encode this in the model.
For example, in the scenario illustrated in Figure~\ref{fig:premon} the agent at $A$ generates a plan to collect the bike at $D$. The solid pink line indicates a typical plan from $A$ to $B$ and then on to $D$. At location $B$ an instruction such as `Go to the house' is clearly ambiguous, potentially leading to confusion.
We are interested in examining whether the user response is different when their route has fewer ambiguous instructions (e.g., the route through $C$).


In order to generate the plans we extend the agent's problem model to incorporate knowledge of the ambiguous instructions. 
This takes the form of a preference that actions that represent ambiguous instructions are avoided (or penalised). Of course, these are soft constraints and in many cases it might be necessary, or expedient to use these actions. To this end we modify the action cost model, manipulating it in order that instructions that are more likely to lead to confusion incur a higher cost.
For each action, $a$, we define the new cost as:
\begin{equation}\label{eq:sim}
cost(a)^\prime = \delta |\{b:\texttt{similar}(a,b)\}| + cost(a)
\end{equation}
For the purposes of this investigation we assume a linear relationship between the number of ambiguous options and the incurred cost, controlled by the parameter $\delta$. 
Notice we are also exploiting the fact that $\forall (a,b): \texttt{similar}(a,b) \implies [\forall s \; s\models a \iff s\models b]$. 
This holds because all groups of actions that can be ambiguous instructions in this domain have the same preconditions.

In the example in Figure~\ref{fig:premon}, the new cost function will penalise the ambiguous instruction in the plan indicated by the solid pink line at $B$. 
The alternative plan through $C$ is not penalised and therefore becomes more attractive.

\begin{figure}
    \centering
    \resizebox{0.4\textwidth}{!}{
    \includegraphics{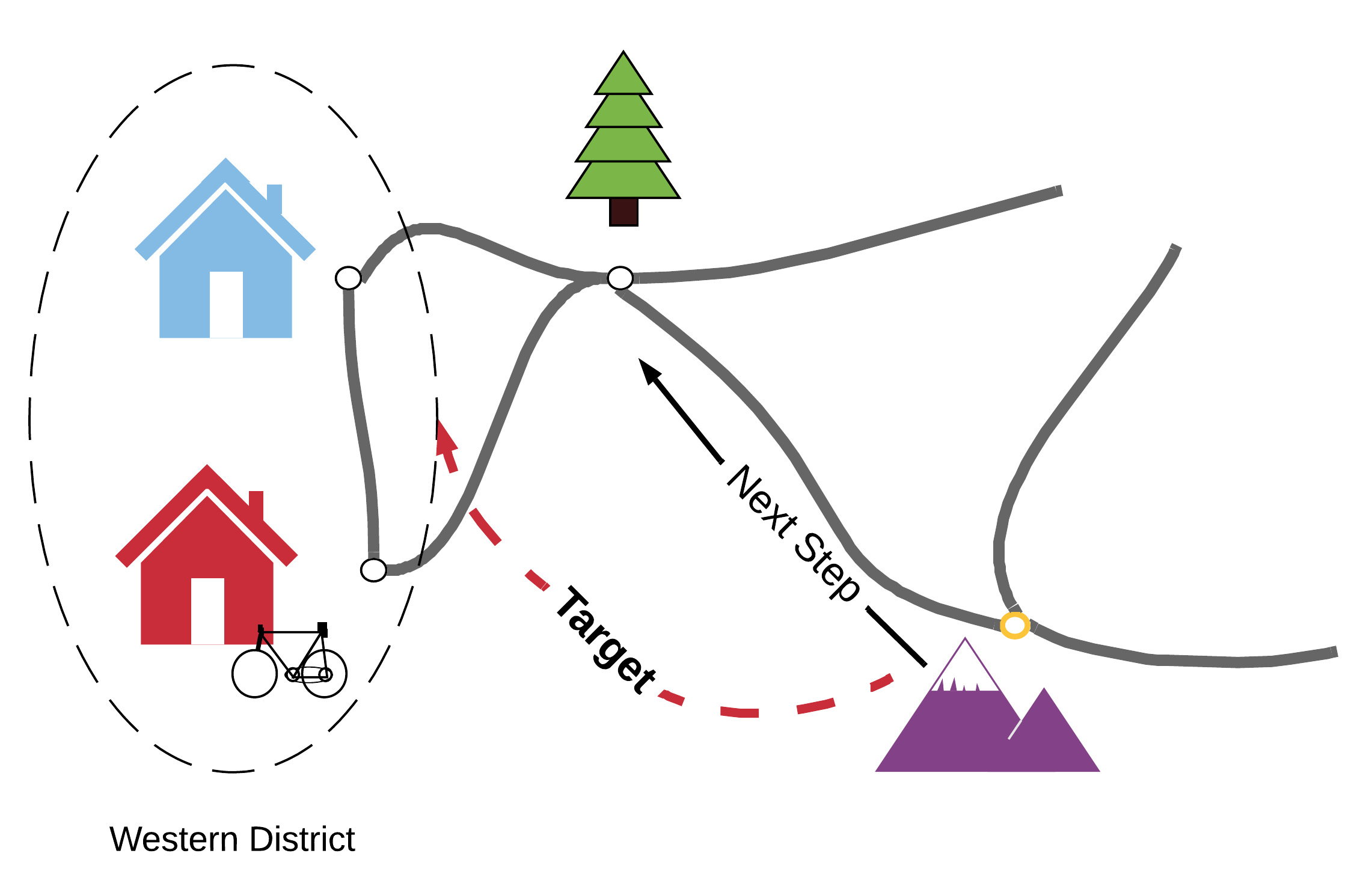}}
    \caption{An example of the agent's intentions for the next step and the motivating target.}
    \label{fig:target}
\end{figure}

\subsection{Plan Summarising}
\label{ssec:target}
Previous results indicate that human interaction partners typically prefer to 
be informed of the plan in advance~\cite{Ganesan_IEEE_2018,bartie2018dialogue}.
As the agent's intentions are captured in its plan, providing a view of the plan to the user allows them to better understand the agent's intentions~\cite{amir2019summarizing}.
In this work, we use the next \emph{target} in the plan to provide a localised view of the plan.
We define \emph{target} as any knowledge gain that reduces the uncertainty in the state (e.g., cancelling out a potential location for a bike) or any action that achieves a subgoal (where that subgoal is not subsequently removed in the plan).
For example, in Figure~\ref{fig:target} an agent at the mountain might have a plan to move to the tree and then on to one of the houses. In this example, the identified target is knowledge about the Western District bike. Whereas the individual plan steps are used for selecting the instructions of the agent, the next target can provide justification, which can explain why the agent is taking an individual step.
We discuss the use of plan summarising further in `Presentation Modes.'

\subsection{Local Inefficiency}

Local inefficiency is where the part of the plan that is being executed could have been replaced by a more efficienct alternative sequence, relative to some cost function.
\begin{definition}
For a plan trace, $\tau=s_0,a_1,\dots,a_n,s_n$, the subsequence $a_i,\dots,a_j$ is locally inefficient 
if
$\exists a_1^\prime,\dots,a_m^\prime: s_j=a_m^\prime(..a_1^\prime(s_{i-1})..) \wedge cost(a_1^\prime,\dots,a_m^\prime)< cost(a_i,\dots,a_j)$.
\end{definition}
For a state, $s_x$, and look-back and look-ahead parameters: $(\theta^<,\theta^>)$, we can perform a local bounded inefficiency analysis for the plan fragments in the interval, $x-\theta^<$ to $x+\theta^>$. 

The reason for perceived local inefficiency may be a deliberate choice, lack of planning resource (i.e., this fragment is only part of a much longer plan) or through alternative human preference.
Consider again the example in Figure~\ref{fig:premon} and if we assume that the route through $C$ is clearly longer. The agent's decision to choose the path through $C$, in order to reduce the number of confusing instructions, is not necessarily clear to the user. The agent might therefore explain its routing choice, e.g., `I chose this route because it was easier to explain. It is only a little longer.'.
We expect that as human preferences are better understood, similar analysis using a human inspired cost function will allow a deeper analysis of perceived inefficiency.

\subsection{Presentation Modes}
The pathway towards generating effective explanations must balance the agent's (lack of) knowledge with human preference's for receiving and giving information.
While there is evidence that humans will typically underexplain and only provide explanations when it becomes clear that there is a problem~\cite{anderson1991hcrc}, it has also been shown that humans can perform better when they are informed of what is going on in advance \cite{foster2009evaluating}.
Understanding how these competing aspects relate to our setting will allow us to better manage the agent's information presentation.

In the `Plan Summarising' subsection the current instruction was explained in the context of the agent's current target. This explanation can be used as a pre- or post-explanation. As a post-explanation the explanation can be used as a reaction to the listener appearing confused or uncertain. 
For example, if the listener in Figure~\ref{fig:target} believed they were heading east and thinks the agent has made a mistake then
indicating the agent's current intention would help.
The explanation can also be used as a pre-explanation, so that the user is given the justification of the following subsequence of actions in advance.
The comparison of these cases: informed of the agent's intentions or not, will be used to examine the difference in user behaviour when encountering ambiguous instructions.
In~\cite{foster2009evaluating}, they demonstrated that this was beneficial in a joint task that was cleanly separated into stages. In our task some targets only reflect partial progress (e.g., discovering a bike is not at a location) meaning we can observe whether pre-explanations of this sort would still be beneficial when the tasks are interleaved.

\subsection{Initiative Switch}
To simulate a situation where the user knows more than the system, we have made some bikes visible to the user from the surrounding landmarks.
In these situations the agent will still be unaware of the bike's actual position.
We are interested in observing the user response when their knowledge leads them to believe that the agent's instruction is not effective.
We will examine whether offering an initiative switch, so that the user can exploit their knowledge, is preferred by the user and whether it leads to different behaviour.

\section{The Agents}
To gain a deeper understanding of the relationship between a human's reaction and their knowledge it is important that we also control what and when information content is provided to the human.
As a result, we have designed two alternative agents: a responsive agent (Agent 1) and a partially predictive agent (Agent 2). Both agents are defined using a shared behaviour template, which makes for an easier comparison between the conditions. 
Due to the online nature of the study and the associated restrictions an agent can only respond to user inaction (not dialogue or more informative signals). 
As such, in some cases we allow the agents to use certain information that might otherwise require dialogue to be obtained in a real environment. 



\begin{table}
\centering
\begin{tabular}{r|l}
\bf Dialogue Action 1 & e.g., `Move to the House'\\
\bf User Action Timer &Wait: Timer 1:  \\
\bf Dialogue Action 2 & e.g., `The Blue one'\\
\bf User Action Timer &Wait: Timer 2 \\
\bf Repeat &While no user action
\end{tabular}
\caption{\label{tab:sbeh} A behavioural template with space for two main dialogue actions, e.g., an initial instruction and then further elaboration or an explanation.}
\end{table}


\subsection{The Behaviour Template}
We have defined a simple behaviour template to underpin the agents' general behaviour.
A simplified template for the example in Figure~\ref{fig:amb} is presented in Table~\ref{tab:sbeh}. 
The agent's dialogue is defined in two main parts: 
an initial dialogue action (e.g., for an instruction) and a second dialogue action, which occurs after user hesitation (e.g., for elaboration or explanation). 
Although not shown, our template also distinguishes between landmarks that have previously been visited (to allow an alternative dialogue action, such as `Go back to the house') and slots for transition based actions (e.g., a \texttt{pickup-bike} action can be linked with an acknowledgement).
A key aspect of the behaviour templates in the context of the user study is a set of shared parameters (e.g., timing) that define how the templates are mapped onto behaviours.

  
      

\begin{table}
\centering
\begin{tabular}{r|l}
\bf Dialogue Action 1 & `Move to the Cafe'\\
\bf User Action Timer &Wait: 2 seconds  \\
\bf Dialogue Action 2 & `Next is the Eastern bike.'\\
\bf User Action Timer &Wait: 5 seconds \\
\bf Repeat &While no user action
\end{tabular}
\caption{\label{tab:tempage1} An example template for Agent 1. The next target is used in dialogue action 2 to justify moving to the House.}
\end{table}

\newcommand{\pushcode}[1][1]{\hskip\dimexpr#1\algorithmicindent\relax}
\begin{agent}[t]
\caption{{\sc behaviour description}\label{age:1}}
\begin{algorithmic}[1]
\Function{DialogueAction1}{$s,\pi^<,\pi^>=a_0,..,a_n$}
  \If{\texttt{isa}($a_0,move$)}
    \State $\gamma\gets $`Go to the ' + \texttt{type}($a_0$.dest)
  \EndIf
  \State ...
\EndFunction
\Function{DialogueAction2}{$s,\pi^<,\pi^>=a_0,..,a_n$}
  \If{\texttt{isa}($a_0,move$)}
    \If{$\exists op^\prime\, s\models op' \wedge a_0.dest==op^\prime.dest$}
      \State $\gamma\gets $`Go to the ' + \texttt{color}($a_0$.dest) + ` one'
    \Else
      \If{\texttt{LBInefficiency}($\pi^<,\pi^>$)}
        \State $\gamma\gets $`I had a long plan to make...'
      \EndIf
      \State $\gamma\gets\gamma+$`Next is the'+\texttt{target}($\pi^>$)+`bike'
    \EndIf
    
  \EndIf
  \State ...
\EndFunction
\end{algorithmic}
\end{agent}

\subsection{Agent 1: The Responsive Agent}
During the early interactions between a human user and an agent it might be unclear how the agent can best manage the interaction to prevent the user from becoming uncertain or confused.
This contrasts directly with the assumption typically made in model reconciliation-based explanation approaches~\cite{chakraborti2017plan}, where both the human's model of the environment and their reasoning capabilities are accessible.
We are investigating how human communication signals can be used as an indication of the user's requirements, e.g., such as a need of more information.

The first agent's behaviour is defined by the rules indicated in the behaviour description in Agent~\ref{age:1}. 
These rules are used to parameterise the template, presented in Table~\ref{tab:sbeh}.
For example, Table~\ref{tab:tempage1} instantiates the template for the agent's strategy for the example in Figure~\ref{fig:premon}.
Each episode starts with the instruction of where to go, or what to do. 
We will focus on move actions (e.g., lines $2$ and $8$), though other actions are mapped in a similar way.
For a move action the agent uses the landmark type (e.g., house) to indicate the next landmark (line $3$).
In the case of user inaction then the agent uses its second dialogue action.
If the instruction is ambiguous then they use this in order to elaborate on the instruction (line $10$). 
Otherwise the agent indicates their intended target (line $15$), as described above and in the case that their plan is locally inefficient then they precede the target with an acknowledgement of the inefficiency (line $13$).
The agent also implements default acknowledgement actions, e.g., when a bike is picked up or the user goes the wrong way.

\begin{table}
\centering
\begin{tabular}{r|l}
\bf Dialogue Action 1 & `Next is the Eastern bike.' \\
&`Move to the Tree'\\
\bf User Action Timer &Wait: 2 seconds  \\
\bf Dialogue Action 2 & `I chose this route because\\
{}&it was easier to explain..'\\
\bf User Action Timer &Wait: 5 seconds \\
\bf Repeat &While no user action
\end{tabular}
\caption{\label{tab:tempage2} An example template for Agent 2, where the agent informs the listener of the next target in dialogue action 1 and then justifies its choice of a longer route.}
\end{table}

\begin{agent}[t]
\caption{{\sc behaviour description}\label{age:2}}
\begin{algorithmic}[1]
\Function{DialogueAction1}{$s,\pi^<=..a_{-1},\pi^>=a_0..$}
  \If{\texttt{isa}($a_0,move$)}
    \If{$\lnot\texttt{target}(a_{-1},\pi^>)==\texttt{target}(\pi^>)$}
      \State $\gamma\gets $`Next is the '+\texttt{target}($\pi^>$)+` bike'
    \EndIf
    \State $\gamma\gets\gamma+$`Go to the ' + \texttt{type}($a_0$.dest)
  \EndIf
  \State ...
\EndFunction
\Function{DialogueAction2}{$s,\pi^<,\pi^>=a_0,..,a_n$}
  \If{\texttt{isa}($a_0,move$)}
    \If{$\exists op^\prime\, s\models op' \wedge a_0.dest==op^\prime.dest$}
      \State $\gamma\gets $`Go to the ' + \texttt{color}($a_0$.dest) + ` one'
    \Else
      \If{\texttt{K\_u\_shortcut}($s,a_0$)}
        \State $\gamma\gets $`Can you see it? Go ahead...'
      \ElsIf{\texttt{LBInefficiency}($\pi^<,\pi^>$)}
        \State $\gamma\gets $`I chose this route because...'
      \EndIf
      \State $\gamma\gets\gamma+$`Next is the'+\texttt{target}($\pi^>$)+`bike'
    \EndIf
    
  \EndIf
  \State ...
\EndFunction
\end{algorithmic}
\end{agent}

\subsection{Agent 2: The Predictive Agent}
In order to allow a deeper analysis of human response and preferences we have also developed a second, partially-predictive, agent.
This agent combines a strategy of pre-explaining with an exploitation of a richer understanding of the domain in order to reduce the user's uncertainty through both avoiding (forward planning) and preparation (providing information in advance).
The agent's planning model captures knowledge of ambiguous instructions in its cost model, as presented in the `Explicable Plans' subsection. The generated plans will therefore tend to lead to sequences where the user will be more certain of the intention of the instruction.

The agent's behaviour is given in the behaviour description in Agent~\ref{age:2}. 
In this case, if the agent has just changed to focus on a new target since the previous action (e.g., line $3$) then the new target is indicated to the user (line $4$).
This is illustrated in Table~\ref{tab:tempage2}, which instantiates the behaviour template for the agent's strategy for the example in Figure~\ref{fig:premon}.
The agent (also) presents the next instruction as before.
A brief (one step) initiative switch is offered (line 16) in the case where the agent is indicating a possible bike position, but there is an alternative applicable action that leads to a bike and the user can see the bike.
The agent's plan is generated with the knowledge of the ambiguous instructions encoded in its cost model. However, we use the original model to measure inefficiency, as the user will not have the knowledge to consider the instruction complexity. 
Therefore the justification for the inefficient plan (assuming a good enough plan is obtained) is attributed to the agent's motivation to simplify the instructions (line $18$), as mentioned above.

\begin{figure}
    \centering
    \resizebox{0.45\textwidth}{!}{
    \includegraphics{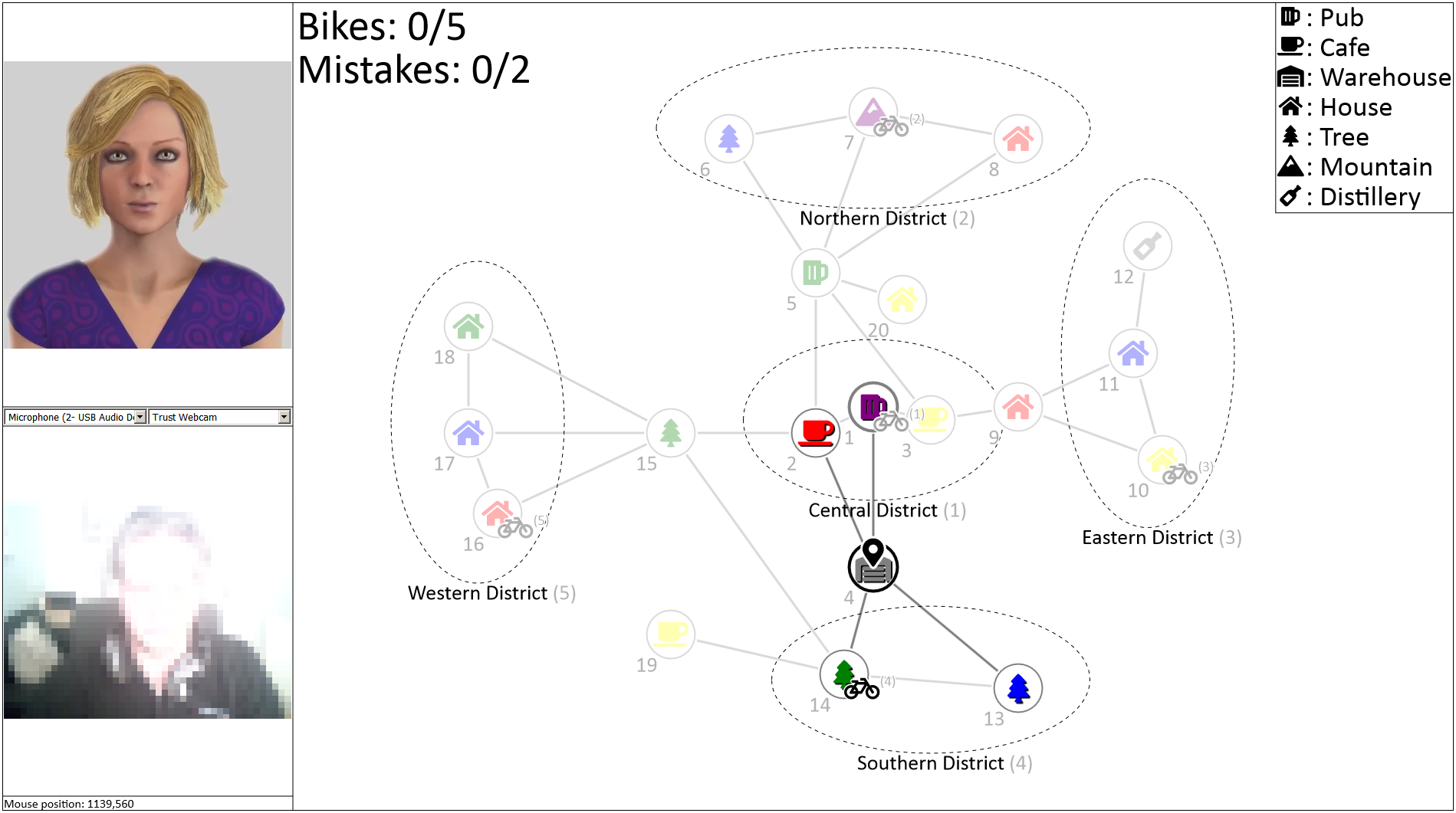}}
    \caption{A screenshot of the system used in our study. The user is able to see where the districts are positioned, as well as the nearby landmarks (note: greyed out landmarks and roads would not be presented to a participant). Their view includes a virtual instruction giver (top left), which is used for interaction and their own video feed (bottom left).}
    \label{fig:screenshot}
\end{figure}

\section{The User Study}
We are currently preparing to conduct an online user study to investigate the following questions:
\begin{enumerate}
\item Do unexpected situations/instructions lead to detectable (i.e., biometric, behavioural or linguistic) user response?
\item And if yes: does the reason why the situation/instruction is unexpected lead to different types of user response?
\item Does the user's response change as the task progresses?
\item Does providing the user with explanations lead to improved confidence and trust in the system?
\item Can simple user heuristics be used to explain aspects of the user's behaviour (nearest, least visited, intent)?
\end{enumerate}
These questions are aimed at deepening our understanding of how planners can be used to better manage human interactions.
We aim to investigate human responses, firstly to identify predictors so that we can provide more information to the planner during the interaction, providing the opportunity to respond to the situation reactively. More specifically we are investigating whether different situations lead to different responses, e.g., uncertainty about instruction ambiguity or confusion over strategy. This finer granularity will provide more information to the planner, allowing it to make more informed choices in future interactions.
We are also investigating default behaviours and preferences, which will inform on the selection of explanation strategy, e.g., using pre- or post-explanations, and potentially informing the way plans are constructed.
As well as gathering objective data (as described below) we will also collect subjective data from surveys. This will allow us to better understand the user's experiences and preferences, and enable us to analyse their perception of the virtual humans.
In the remainder of this section we will discuss how the interaction is controlled to ensure the intended user experience and then provide an overview of the system that we have developed to support the online study.

\subsection{Controlling the Interaction}
The first three questions require that we can observe user response in a variety of situations and that the user receives the appropriate information in each situation for each condition.
The agent is a key component for controlling the information that is presented to the listener and their rule-based behaviour descriptions (see `The Agents' Section) allow their response to be conditioned on the current situation.
We will use four conditions: the responsive and predictive agents described above and two baseline agents, each providing different information or a different strategy at different stages. This will allow us to observe how the user's behaviour and response is altered by the strategy being used and the information they have been provided.

The map design is also an important aspect for ensuring appropriate situations are experienced during the interaction, including ambiguity and knowledge difference. For example, we showed in Figure~\ref{fig:amb} how placing similar landmarks at adjacent positions can lead to ambiguous instructions. 
Ambiguous instructions are important as they allow us to observe human reaction to uncertainty: how they respond and what they choose (e.g., do they pick the closest one or the one they have not already visited?).
This is particularly interesting in combination with the information provided by the agent, e.g., is behaviour different when the user already knows the agent's intentions?
The maps therefore have been carefully designed to ensure that execution sequences will encounter these situations.



\subsection{The Study Website}
A screenshot from our system is presented in Figure~\ref{fig:screenshot} (see~\cite{lindsaySMSKEPS2020} for more details), with the map in the centre and the virtual human and 
user feed presented on the left. 
The virtual instruction giver provides the user with verbal instructions, such as `Go to the house.' 
The user feed informs the participant of the major information/data that is being recorded (e.g., video and audio).
The main portion of the screen is taken up by the interactive map on which the experiment plays out.

The back-end server-based part of the system collects and aggregates the behavioural and interaction data from the participant for later offline study and analysis.
The recorded data includes video (for facial expression analysis), audio and mouse movement data, as well as in task events, such as recording the user's choices and their execution trace through the task.
All data collected is time-stamped, and aggregated for an individual user,
 allowing a full analysis of their interaction with the system.


\begin{table*}
\centering
\resizebox{0.7\textwidth}{!}{
\begin{tabular}{ r||l|l|lll}
\hline 
\multirow{2}{*}{Explanation$\backslash$Planner} &\multicolumn{2}{c|}{Reactive} &\multicolumn{3}{c}{Predictive}\\
    {}  &R-FF &R-Lama &$\delta=1$ &$\delta=2$ &$\delta=3$ \\
\hline 
$|\pi|$ &35.82 (2.10) &28.40 (1.88) &27.36 (1.63) &27.78 (1.78)  &28.72 (1.77)  \\
$|similars|$ &16.83 (1.51) &12.54 (1.22) &9.92 (1.08) &9.54 (0.92) &9.60 (0.94)  \\
\hline
Move &30.60 (1.99) &23.18 (1.74) &22.13 (1.49) &22.57 (1.66) &23.48 (1.68)  \\
Pickup &5.22 (0.43) &5.22 (0.46) &5.23 (0.46) &5.21 (0.44) &5.24 (0.46)  \\
\hline
Elaborate &14.82 (1.38) &11.47 (1.20) &8.90 (1.02) &8.52 (0.87) &8.56 (0.89)  \\
PreTarget(K) &0.00 (0.00) &0.00 (0.00) &5.46 (0.92) &6.43 (0.92) &8.05 (0.57)  \\
Target(K) &13.66 (1.49) &10.62 (0.98) &12.17 (1.29) &12.92 (1.13) &13.77 (1.19)   \\
Target(Pos) &2.12 (0.35) &1.09 (0.38) &1.06 (0.24) &1.13 (0.45) &1.15 (0.50)  \\
Inefficient &1.11 (0.50) &0.00 (0.00) &0.03 (0.18) &1.04 (0.50) &1.05 (0.51)  \\
Initiative &0.00 (0.00) &0.00 (0.00) &0.97 (0.32) &1.03 (0.21) &1.04 (0.22) \\
\hline 
\end{tabular}}
\caption{\label{tab:exp} $1000$ executions were simulated with $\gamma=0.95$ for Map 1 and policies generated using alternative classical planning approaches. The means (and standard deviations) of the sequence length, number of alternatives compatible with the instruction ($|similars|$) and number of instructions and explanations encountered were recorded. }
\end{table*}

\section{Empirical Analysis}
As part of our preparations for conducting the online user study we have tested our approach in order to examine the behaviours of the reactive and predictive agents. 
In this section, we present empirical results generated as part of this testing.
Our approach to partially-observable planning uses K-Replanner~\cite{bonet2011planning}, which supports efficient plan generation for partially-observable problems via replanning.
K-Replanner exploits a compilation of the problem to classical planning and we have analysed using alternative planners/configurations to understand the expected behaviour of the resulting agent and the policies generated.

In our user study, each participant will be presented with four conditions. As such, we constructed four maps each with around twenty landmarks and five bikes (similar to the graph in Figure~\ref{fig:screenshot}). The goal of each problem was to find and collect the bikes and return to the base. 
The planners/configurations we used with a responsive  strategy (Agent~1, above) were:
\begin{description}
    \item[\bf R-FF] The FF planning system~\cite{hoffmann2001ff}.
    \item[\bf R-Lama] The LAMA-11 configuration of Fast Downwards~\cite{richter2010lama} with a $3$ second time-out (the average was $1.5$ seconds).
\end{description}
The planners/configurations we used with a predictive strategy (Agent~2, above) were:
\begin{description}
    \item[\bf Predictive ($\delta=\{1,2,3\}$)] Planning model with knowledge of ambiguous instructions. $\delta$ controls balance between avoiding ambiguous questions and plan length (see Equation~\ref{eq:sim}). LAMA-11 is used with a $5$ second time-out.
\end{description}
Each configuration was used within K-Replanner and the resulting plans were converted into state action policies.
Each policy allows a bounded amount of exploration, including up to two errors as well as selected additional alternatives, e.g., where the given instruction is ambiguous.
In this way we have allowed some user flexibility, while still generating reasonable sized policies for the online study.
It should be noted that the comparison of quality between these planners is not intended to be \emph{fair} in a traditional sense, e.g., FF returns its first plan, whereas Lama and Predict are being used as anytime planners. However, the intention is to use a variety of generation approaches to provide a wider context for understanding how these planner configurations would lead to alternative behaviours. 
\begin{table}
\centering
\resizebox{.45\textwidth}{!}{
\begin{tabular}{ l|r||l|l}
     \multicolumn{2}{c||}{Map$\backslash$  Planner}&R-Lama &Predictive ($\delta=1$) \\
\hline 
\multirow{2}{*}{Map1} &$|\pi|$ &28.40 (1.88) &27.36 (1.63) \\
{} &$|similars|$ &12.54 (1.22) &9.92 (1.08) \\
\hline
\multirow{2}{*}{Map2} &$|\pi|$ &25.60 (1.75) &22.69 (1.82) \\
{} &$|similars|$ &11.86 (1.19) &8.89 (1.04)  \\
\hline
\multirow{2}{*}{Map3} &$|\pi|$ &24.55 (1.80) &28.74 (1.91)  \\
{} &$|similars|$ &15.01 (1.70) &10.78 (1.46)  \\
\hline
\multirow{2}{*}{Map4} &$|\pi|$ &32.78 (1.68) &37.04 (2.61) \\
{} &$|similars|$ &13.75 (1.29) &10.38 (1.16) \\
\hline
\end{tabular}}
\caption{\label{tab:sims} Sampled means (and standard deviations) of the execution length and number of similar alternatives for $1000$ policy samples, for the $4$ maps and $\gamma=0.95$.}
\end{table}

\subsection{Generation of Explanations}
In order to compare both the strategies captured by the policies and the explanation generation we used each policy to generate execution samples.
A parameter, $\gamma$, was used to control the probability that the simulated user would follow each instruction. In this experiment this parameter was fixed at $\gamma=0.95$ in all runs.
Table~\ref{tab:exp} presents the executions: the execution details, the encountered instructions and the generated explanations, for Map 1.
The results show the average execution length ($|\pi|$) and the average number of alternatives that were compatible with an instruction ($|similars|$). 
`Elaborate' counts the number of move instructions that were ambiguous, `Inefficient' counts cases of detected local inefficiency and `initiative' counts the number of offers to take the initiative.
The explanation counts cover both dialogue actions (see Table~\ref{tab:sbeh}). For example, in Agent~\ref{age:1} the targets: bike target (`Target(K)') and destination target (`Target(Pos)'), are presented as part of the responsive explanation strategy and only occur in dialogue action 2. Whereas Agent~\ref{age:2} also uses the targets (e.g., `PreTarget') in dialogue action 1, to prepare the user.  

The plan lengths indicate that executions using R-FF were longer and typically involved some local inefficiency.
It is unsurprising that as the execution length increases, so do other features, such as number of ambiguous instructions. 
The execution traces generated by R-Lama were shorter with no identified local inefficiency.

The predictive plan lengths and number of ambiguous instructions are similar to each other. 
However, it is clear that the configurations are generating alternative plans.
The increase in `Target(Pos)' suggests that more switches are being made between subtasks.
The executions for $\delta>1$ are both likely to have local inefficiency, unlike $\delta=1$. 
Notice in the case of the predictive agents, local \emph{inefficiency} is more likely to be due to avoiding ambiguous instructions.

\subsection{Avoiding Uncertain Actions}
Table~\ref{tab:sims} presents the results for plan length and alternatives for the reactive and predictive agents that we will use in the user study on each of the 4 maps. 
It shows that the predictive agent is able to generate plans with fewer ambiguous instructions in each of the maps.
The maps were designed with the aim of putting the participant into ambiguous situations and so it is expected that the predictive agent will still encounter ambiguous instructions.

\section{Related Work}

The increasing adoption of AI Planning in real world applications has led to a growing focus around Explainable Planning (XAIP)~\cite{fox2017explainable}.
Previous work has examined the impact of robot strategy in human-robot interactions:
In ~\cite{Zhang_IROS_2015} they demonstrate that a proactive robot strategy can lead to better team performance, but increase cognitive load; ~\cite{Dragan_acm_2015} examine the related issues of predictability and legibility.

In~\cite{miller2019explanation} it is argued that explainable AI (of which XAIP forms a part) should be based on the findings of previous work in the social sciences, such as cognitive science.
Last year we outlined our intentions to bring together experimental research in cognitive science, involving cooperative joint action, with the practical construction of automated planning tools to apply to the task of explanation generation~\cite{petrickSMS19}.
In~\cite{jobSMS2020} we investigated measures of confidence in the context of an instruction-giving task, where biometric and behavioural measures (eye movements, galvanic skin response, facial expression, and task performance) were recorded. These were analysed in conjunction with subjective, self-report measures of confidence combined with additional perceptions of the virtual human during the interactions. 
The study in this paper aims to clarify and extend these results in the context of plan-based agent interaction (albeit with a reduced scope of user inputs, due to the online nature of the study).

As mentioned above, the task used in this work was inspired by the HCRC Map Task~\cite{anderson1991hcrc} and the GIVE Challenge~\cite{byron2009report}.
In~\cite{koller2011experiences}, the GIVE Challenge task was used to 
evaluate whether classical planning was an effective approach to natural language generation. 
Their work focused on instruction giving and did not consider explanations or managing an interaction.
In~\cite{Petrick-Foster:2013}, they presented a robot bartender that could successfully balance multiple simultaneous customers, using the
knowledge-level PKS planner~\cite{petrick2002knowledge} to construct branched (contingent) plans conditioned on the customer's social states.
Our current work intends to inform future work in these sorts of systems by gaining an understanding of a broader range of human communication signals and how they can be utilised to inform action selection.

\section{Conclusion and Future Work}
We intend to conduct a web-based user study to investigate human response during interaction with a plan-based agent. 
We have designed a simple bike sharing task that supports our investigation of several important aspects of joint task interaction, including knowledge difference and uncertainty.
Our study will compare two alternative agents: a reactive agent, representing an initial interaction, and a more predictive one, representing an agent that has knowledge of previous interactions. 
These agents together with careful map design ensure that the participants will be given the appropriate information and experience the situations intended during their interactions.
We presented an empirical analysis, which examines aggregated execution traces for different planning configurations for each of the conditions that will be used in the study.
This includes results showing that the predictive agent generates plans with fewer uncertain situations than the reactive agent.
After the intended user study has been conducted and the data analysed, we hope to conduct a lab-based study to
gather additional biometric data (e.g., GSR and eye movements), which is not practical during online data collection.
Our aim is to enhance our plan-based agent's world model using our improved understanding of human behaviour, in order to enable the agent to respond reactively to user signals, and
harness the representational benefits of approaches like epistemic planning~\cite{Bolander:2017,petrick2002knowledge} in partially-observable domains.

\section{Acknowledgements}

This work is funded by the UK’s EPSRC Human-Like Computing programme under grant number EP/R031045/1.

\footnotesize
\bibliography{bibliography}

\begin{thebibliography}{}

\bibitem[\protect\citeauthoryear{Amir, Doshi-Velez, and
  Sarne}{2019}]{amir2019summarizing}
Amir, O.; Doshi-Velez, F.; and Sarne, D.
\newblock 2019.
\newblock Summarizing agent strategies.
\newblock {\em Autonomous Agents and Multi-Agent Systems}.

\bibitem[\protect\citeauthoryear{Anderson \bgroup et al\mbox.\egroup
  }{1991}]{anderson1991hcrc}
Anderson, A.~H.; Bader, M.; Bard, E.~G.; Boyle, E.; Doherty, G.; Garrod, S.;
  Isard, S.; Kowtko, J.; McAllister, J.; Miller, J.; et~al.
\newblock 1991.
\newblock The {HCRC} map task corpus.
\newblock {\em Language and speech} 34(4):351--366.

\bibitem[\protect\citeauthoryear{Bartie \bgroup et al\mbox.\egroup
  }{2018}]{bartie2018dialogue}
Bartie, P.; Mackaness, W.; Lemon, O.; Dalmas, T.; Janarthanam, S.; Hill, R.~L.;
  Dickinson, A.; and Liu, X.
\newblock 2018.
\newblock A dialogue based mobile virtual assistant for tourists: The
  {SpaceBook} {P}roject.
\newblock {\em Computers, Environment and Urban Systems} 67:110--123.

\bibitem[\protect\citeauthoryear{Bolander}{2017}]{Bolander:2017}
Bolander, T.
\newblock 2017.
\newblock {A Gentle Introduction to Epistemic Planning: The DEL Approach}.
\newblock In {\em Proceedings of the 9th Workshop on Methods for Modalities},
  1--22.

\bibitem[\protect\citeauthoryear{Bonet and Geffner}{2011}]{bonet2011planning}
Bonet, B., and Geffner, H.
\newblock 2011.
\newblock Planning under partial observability by classical replanning: Theory
  and experiments.
\newblock In {\em Proceedings of the International Joint Conference on
  Artificial Intelligence}.

\bibitem[\protect\citeauthoryear{Byron \bgroup et al\mbox.\egroup
  }{2009}]{byron2009report}
Byron, D.; Koller, A.; Striegnitz, K.; Cassell, J.; Dale, R.; Moore, J.~D.; and
  Oberlander, J.
\newblock 2009.
\newblock Report on the first {NLG} challenge on generating instructions in
  virtual environments ({GIVE}).
\newblock In {\em Proceedings of the 12th European Workshop on Natural Language
  Generation}.

\bibitem[\protect\citeauthoryear{Chakraborti \bgroup et al\mbox.\egroup
  }{2017}]{chakraborti2017plan}
Chakraborti, T.; Sreedharan, S.; Zhang, Y.; and Kambhampati, S.
\newblock 2017.
\newblock Plan explanations as model reconciliation: moving beyond explanation
  as soliloquy.
\newblock In {\em Proceedings of the 26th International Joint Conference on
  Artificial Intelligence},  156--163.

\bibitem[\protect\citeauthoryear{Dalzel-Job, Hill, and
  Petrick}{2020}]{jobSMS2020}
Dalzel-Job, S.; Hill, R.~L.; and Petrick, R. P.~A.
\newblock 2020.
\newblock Start making sense: Predicting confidence in virtual human
  interactions using biometric signals.
\newblock In {\em Proceedings of Measuring Behavior}.

\bibitem[\protect\citeauthoryear{Dragan \bgroup et al\mbox.\egroup
  }{2015}]{Dragan_acm_2015}
Dragan, A.~D.; Bauman, S.; Forlizzi, J.; and Srinivasa, S.~S.
\newblock 2015.
\newblock Effects of robot motion on human-robot collaboration.
\newblock In {\em Proceedings of the Tenth Annual ACM/IEEE International
  Conference on Human-Robot Interaction}, HRI '15,  51–58.

\bibitem[\protect\citeauthoryear{Foster \bgroup et al\mbox.\egroup
  }{2009}]{foster2009evaluating}
Foster, M.~E.; Giuliani, M.; Isard, A.; Matheson, C.; Oberlander, J.; and
  Knoll, A.
\newblock 2009.
\newblock Evaluating description and reference strategies in a cooperative
  human-robot dialogue system.
\newblock In {\em Proceedings of the International Joint Conference on
  Artificial Intelligence}.

\bibitem[\protect\citeauthoryear{Fox, Long, and
  Magazzeni}{2017}]{fox2017explainable}
Fox, M.; Long, D.; and Magazzeni, D.
\newblock 2017.
\newblock Explainable planning.
\newblock {\em arXiv preprint arXiv:1709.10256}.

\bibitem[\protect\citeauthoryear{Hoffmann and Nebel}{2001}]{hoffmann2001ff}
Hoffmann, J., and Nebel, B.
\newblock 2001.
\newblock The {FF} planning system: Fast plan generation through heuristic
  search.
\newblock {\em Journal of Artificial Intelligence Research} 14:253--302.

\bibitem[\protect\citeauthoryear{{Kalpagam Ganesan} \bgroup et al\mbox.\egroup
  }{2018}]{Ganesan_IEEE_2018}
{Kalpagam Ganesan}, R.; {Rathore}, Y.~K.; {Ross}, H.~M.; and {Ben Amor}, H.
\newblock 2018.
\newblock Better teaming through visual cues: How projecting imagery in a
  workspace can improve human-robot collaboration.
\newblock {\em IEEE Robotics Automation Magazine} 25(2):59--71.

\bibitem[\protect\citeauthoryear{Koller and
  Petrick}{2011}]{koller2011experiences}
Koller, A., and Petrick, R. P.~A.
\newblock 2011.
\newblock Experiences with planning for natural language generation.
\newblock {\em Computational Intelligence} 27(1):23--40.

\bibitem[\protect\citeauthoryear{Lindsay \bgroup et al\mbox.\egroup
  }{2020}]{lindsaySMSKEPS2020}
Lindsay, A.; Craenen, B.; Dalzel-Job, S.; Hill, R.~L.; and Petrick, R. P.~A.
\newblock 2020.
\newblock Supporting an online investigation of user interaction with an {XAIP}
  agent.
\newblock In {\em {ICAPS 2020} Workshop on Knowledge Engineering for Planning
  and Scheduling (KEPS)}.

\bibitem[\protect\citeauthoryear{Miller}{2019}]{miller2019explanation}
Miller, T.
\newblock 2019.
\newblock Explanation in artificial intelligence: Insights from the social
  sciences.
\newblock {\em Artificial Intelligence} 267:1--38.

\bibitem[\protect\citeauthoryear{Petrick and
  Bacchus}{2002}]{petrick2002knowledge}
Petrick, R. P.~A., and Bacchus, F.
\newblock 2002.
\newblock A knowledge-based approach to planning with incomplete information
  and sensing.
\newblock In {\em Proceedings of the International Conference on {AI} Planning
  and Scheduling},  212--222.

\bibitem[\protect\citeauthoryear{Petrick and
  Foster}{2013}]{Petrick-Foster:2013}
Petrick, R. P.~A., and Foster, M.~E.
\newblock 2013.
\newblock Planning for social interaction in a robot bartender domain.
\newblock In {\em Proceedings of the International Conference on Automated
  Planning and Scheduling}.

\bibitem[\protect\citeauthoryear{Petrick, Dalzel-Job, and
  Hill}{2019}]{petrickSMS19}
Petrick, R. P.~A.; Dalzel-Job, S.; and Hill, R.~L.
\newblock 2019.
\newblock Combining cognitive and affective measures with epistemic planning
  for explanation generation.
\newblock In {\em {ICAPS 2019} Workshop on Explainable Planning (XAIP)},
  141--145.

\bibitem[\protect\citeauthoryear{Richter and Westphal}{2010}]{richter2010lama}
Richter, S., and Westphal, M.
\newblock 2010.
\newblock The lama planner: Guiding cost-based anytime planning with landmarks.
\newblock {\em Journal of Artificial Intelligence Research} 39:127--177.

\bibitem[\protect\citeauthoryear{Zhang \bgroup et al\mbox.\egroup
  }{2015}]{Zhang_IROS_2015}
Zhang, Y.; Narayanan, V.; Chakraborti, T.; and Kambhampati, S.
\newblock 2015.
\newblock A human factors analysis of proactive support in human-robot teaming.
\newblock In {\em IEEE International Conference on Intelligent Robots and
  Systems}.

\end{thebibliography}
\bibliographystyle{aaai}
\end{document}